\documentclass[runningheads]{llncs}
\usepackage{caption}
\usepackage{subcaption}
\usepackage[T1]{fontenc}
\usepackage{graphicx}
\usepackage{comment}
\usepackage{bbding}

\usepackage{url}            
\usepackage{multirow}
\newcommand{\nop}[1]{}

\usepackage{mathabx}

\newcommand\doubleplus{+\kern-1.3ex+\kern0.8ex}
\newcommand{\serie}[2]{#1\smalltriangleright#2}
\newcommand{\dserie}[2]{\serie{#1-#2}{#1}}
\newcommand{\ser}[4]{#1^#2_{\dserie{#3}{#4}}}
\newcommand{\cD}{\textrm{Conv}^{\downarrow}}
\newcommand{\cU}{\textrm{Conv}^{\uparrow}}

\usepackage{orcidlink}

\begin{document}
\title{Temporal Saliency Detection Towards Explainable Transformer-based Timeseries Forecasting}
\titlerunning{Temporal Saliency Detection}
%
\author{Nghia Duong-Trung\inst{1}\Envelope \orcidlink{0000-0002-7402-4166} \and
Duc-Manh Nguyen\inst{2}\orcidlink{0000-0002-9668-4974} \and
Danh Le-Phuoc\inst{2}\orcidlink{0000-0003-2480-9261}}
\authorrunning{Duong-Trung, Nguyen, Le-Phuoc}
%
\institute{German Research Center for Artificial Intelligence (DFKI) \\
	 Alt-Moabit 91 C, 10559 Berlin, Germany. \\ \email{nghia\_trung.duong@dfki.de}\\ 
\and
Technische Universität Berlin \\Straße des 17. Juni 135, 10623 Berlin, Germany
\email{duc.manh.nguyen@tu-berlin.de,danh.lephuoc@tu-berlin.de}
} 
\maketitle              
\begin{abstract}


Despite the notable advancements in numerous Transformer-based models, the task of long multi-horizon time series forecasting remains a persistent challenge, especially towards explainability. 
Focusing on commonly used saliency maps in explaining DNN in general, our quest is to build attention-based architecture that can automatically encode saliency-related temporal patterns by establishing connections with appropriate attention heads. 
Hence, this paper introduces Temporal Saliency Detection (TSD), an effective approach that builds upon the attention mechanism and applies it to multi-horizon time series prediction. 
While our proposed architecture adheres to the general encoder-decoder structure, it undergoes a significant renovation in the encoder component, wherein we incorporate a series of information contracting and expanding blocks inspired by the U-Net style architecture. 
The TSD approach facilitates the multiresolution analysis of saliency patterns by condensing multi-heads, thereby progressively enhancing the forecasting of complex time series data. 
Empirical evaluations illustrate the superiority of our proposed approach compared to other models across multiple standard benchmark datasets in diverse far-horizon forecasting settings. 
The initial TSD achieves substantial relative improvements of 31\% and 46\% over several models in the context of multivariate and univariate prediction. 
We believe the comprehensive investigations presented in this study will offer valuable insights and benefits to future research endeavors.

\end{abstract}
\section{Introduction}
\label{sec:introduction}

Time series forecasting empowers decision-making on chronological data, and performs an essential role in various research and industry fields such as healthcare \cite{piccialli2021artificial}, energy management \cite{stefenon2022time}, industrial automation \cite{yang2021novel}, planning for infrastructure construction \cite{sartirana2022data}, economics and finance \cite{sezer2020financial}. 
Time series observations can be a single sequence addressed by traditional time series forecasting approaches such as autoregressive integrated or exponentially weighted moving averages.
However, actual time series data may consist of several channels as predictors for future forecasting and thus require more effective approaches.
Multi-horizon prediction allows us to estimate a long sequence, optimizing intervened actions at multiple time steps in the future where performance improvements are precious.   
Hence, a significant challenge for time series forecasting is to develop practical models dealing with the heterogeneity of multi-channel time series data and produce accurate predictions in multi-horizon.

\begin{figure}[ht!]
	\centering
	\includegraphics[width=0.99\columnwidth]{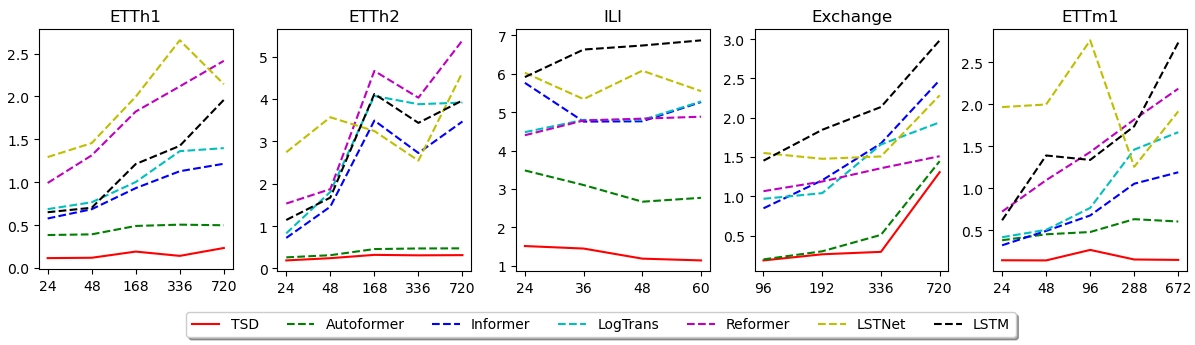} %
	\caption{Multivariate time series forecasting comparison. The lower lines indicate better forecasting capacity. \textit{y}-axis indicates the MSE loss while \textit{x}-axis presents the prediction horizon.}
	\label{fig:plot-comparison}
\end{figure}

Following the great success of attention mechanisms in machine translation, recent research has adapted it for time series forecasting \cite{lim2021temporal,zhou2021informer,zeng2021topological,gao2022self,jin2022domain,zeng2023transformers}.
Self-attentions consider the local information that the models only utilize point-wise dependencies.
Benefiting from the self-attention mechanism, Transformers achieve significant efficiencies in dependency modeling for sequential data, allowing for constructing more powerful large models. 
However, the forecasting problem is highly challenging in the long term and when many variables affect the target. 
First, detecting temporal dependencies directly from long-horizon time series is unreliable because the dependencies can be spread across many variables, and each variable tends to be different. 
Second, the canonical Transformer with self-attention requires high-power computing for long-term forecasting because of the quadratic complexity of the sequence length. 
Thus, to solve the computational hardware bottleneck, the previous Transformer-based predictive models mainly focused on improving the full self-attention to a sparse version. 
For the information aggregation, Autoformer adopts the time delay block to aggregate the similar sub-series from underlying periods rather than selecting scattered points by dot-product. 
The Auto-Correlation mechanism can simultaneously benefit the computation efficiency and information utilization from the inherent sparsity and sub-series-level representation aggregation.
Despite the significantly improved performance, this approach has to sacrifice information usage because point connections lead to long-term time series forecasting bottlenecks that make it difficult to explain the trained models. 


Towards explainability, saliency maps have been widely used to highlight important input features in model predictions to explain model behaviors in computer vision \cite{yun2022panoramic,morrison2023shared,hu2023xaitk} and recently also apply for time series. 
In fact, \cite{NEURIPS2021_e0cd3f16} introduces the saliency-guided architecture that shows it works CNN and LSTM. 
However, it is still being determined whether such an approach will work with the recent transformer-based approach above, especially with the ones using sparsifying techniques. 
For example,~\cite{Ismail:2020} pointed out that attention-based methods can be insufficient for interpreting multivariate time-series data, e.g., saliency maps fail to reliably and accurately identify feature importance over time in time series data.
We hypothesize that we need a technique to automatically encode the \emph{saliency-related temporal patterns} via connecting to the suitable attention heads.
Thus, we attempt to go beyond heuristic approaches such as Informer, e.g., sparse point-wise connection, Autoformer, e.g., sub-level-wise connection, and propose a generic architecture to empower forecasting models with automatic segment-wise interpolation. Inspired by saliency detection theory \cite{zhang2018review} in images and video recognition \cite{ullah2020brief,li2019motion}, we propose a method to weigh the proper attention to possible emerging temporal patterns.

This method is realized as a learning architecture called Temporal Saliency Detection, which can be categorized into Transformer-family for far-horizon time series forecasting.
Our proposed architecture still follows a general encoder-decoder structure but renovates the encoder component with a series of information contracting and expanding blocks inspired by U-Net style architecture~\cite{ronneberger2015u}.
The fundamental architecture of a U-net model comprises two distinct paths. 
The first path, referred to as the contracting path, encoder, or analysis path, closely resembles a conventional convolutional network and is responsible for extracting pertinent information from the input data. 
On the other hand, the second path, known as the expansion path, decoder, or synthesis path, encompasses up-convolutions and feature concatenations derived from the contracting path. 
This expansion process facilitates learning localized extracted information and concurrently enhances the output resolution. 
Subsequently, the augmented output passes through a final convolutional layer to generate the fully synthesized data. The resultant network exhibits an almost symmetrical configuration, endowing it with a U-like shape.
To this end, similar to the latent diffusion models for high-resolution image synthesis \cite{DBLP:journals/corr/abs-2112-10752}, by allowing the attention mechanism and the information contraction to work in concert, our architecture can construct temporal saliency patterns through segment-wise aggregation. 
While U-Net style architecture \cite{siddique2021u} can pay the way for temporal saliency patterns to emerge like in~\cite{ijcai2021p397}, the challenge is how to match the performance of heuristics point-wise or sub-series-wise methods ( e.g., Informer, LogTrans, and Autorformer). 
The experiments in Section\ref{sec:experiment} show that our approach can be competitive with several state-of-the-art methods as visually summarized in Figure~\ref{fig:plot-comparison}.
The contributions of the paper are summarized as follows:

\begin{itemize}
	\item We introduce the Temporal Saliency Detection model as a harmonic combination between encoder-decoder structure and U-Net architecture to empower the far-horizon time series forecasting towards explainability.
	\item Our proposed approach discovers and aggregates temporal information at the segment-wise level. TSD consistently performs the prediction capacity in ten different multivariate-forecasting horizons.
    \item At the current initial investigation, TSD achieves a \textbf{31\%} and \textbf{46\%} relative improvement over compared models under multivariate and univariate time series forecasting on standard benchmarks.
\end{itemize}

\section{Related Work}

\textbf{Long and Multi Horizon Time Series Forecasting} has been a well-established research topic with a steadily growing number of publications due to its immense importance for real applications \cite{masini2021machine,meisenbacher2022review}.
Classical methods such as ARIMA \cite{ariyo2014stock}, RNN \cite{wen2017multi,yu2017long,rangapuram2018deep}, LSTM \cite{bahdanau2014neural} and Prophet \cite{taylor2018forecasting} serve as a standard baselines for forecasting.
One of the most ubiquitous approaches in a wide variety of forecasting systems is deep time series which have been proven effective in both industries \cite{olivares2021probabilistic,makridakis2021predicting} and academic \cite{hewamalage2021recurrent,tealab2018time}.
Amazon time series forecasting services build around DeepAR \cite{salinas2020deepar}, which combines RNNs and autoregressive sliding methods to model the probabilistic future time points.
Attention-based RNNs approaches capture temporal dependency for short and long term predictions \cite{qin2017dual,shih2019temporal,song2018attend}.
CNN's models for time series forecasting also provide a noticeable solution for periodically high-dimensional time series \cite{wang2019multiple,lara2021experimental,lai2018modeling}. 
Another deep stack of fully-connected layers based on backward and forward residual links, named N-BEATS, was proposed by \cite{oreshkin2019n} and later improved by \cite{challu2022n}, called N-HITS, have empowered this research direction. 
Those forecasting approaches focus on temporal dependency modeling by current knowledge, recurrent connections, or temporal convolution.

\textbf{Transformers Based on the Self-attention Mechanism}, originating from the machine translation domain, have been successfully adapted to address different time series problems \cite{lai2018modeling,fan2019multi,shih2019temporal,song2018attend,ma2019cdsa,li2019enhancing}.
Attention computation allows direct pair-wise comparison to any uncommon occurrence, e.g., sale seasons, and can model temporal dynamics inherently.
However, pair-wise interactions make attention-based models suffer from the quadratic complexity of sequence length.
Recent research, Reformer \cite{kitaev2020reformer}, Linformer \cite{wang2020linformer}, and Informer \cite{zhou2021informer}, proposes multiple variations of the canonical attention mechanisms have achieved superior forecasting while in parallel reducing the complexity of pair-wise interactions.
Another exciting paper that belongs to the attention-based family of models is the Query Selector \cite{klimek2021long}, where the idea of computing a sparse approximation of an attention matrix is exploited.
Note that these forecasting models still rely on point-wise computation and aggregation. 
Nevertheless, those models have improved the self-attention mechanism from a full to a sparse version by sacrificing information utilization. In this context, we call them as \emph{sparsification architectures}. 
Along the same line, YFormer \cite{madhusudhanan2021yformer} adjusted the Informer model by integrating a U-Net architecture into Informer's \emph{ProbSparse Self-attention} module. 
While it is also inspired by U-Net like us, YFormer inherits the problem of sacrificing the information utilization of Informer.
Originated from the medical image segmentation problem, U-Net is capable of condensing input information to several intermediate embeddings and up-sampling them to the same resolutions as the input \cite{ronneberger2015u,kohl2018probabilistic,ibtehaz2020multiresunet}.
Apart from the image domain, the U-Net approach has proven noticeable results for sequence modeling \cite{stollerseq} and time series segmentation \cite{perslev2019u}.

\textbf{Auto-correlation Mechanism}
A noticeable encoder-decoder architecture that utilizes Fourier transform is Autoformer \cite{xu2021autoformer} with decomposition capacities and an attention approximation.
Autoformer is based on the series periodicity addressed in the stochastic process theory, where trend, seasonal, and other components are blended \cite{duarte2019decomposing}.
Hence, the model does not depend on temporal dependency as with the transformer-based solutions, but the auto-correlation emerging from data.
The series-wise connections replace the point-wise representation.

\textbf{Saliency} has emerged as a prominent and effective method for enhancing interpretability, providing insights into why a trained model produces specific predictions for a given input.
One approach to leveraging saliency is through saliency-guided training, which aims to diminish irrelevant features' influence by reducing the associated gradient values. 
This is achieved by masking input features with low gradients and then minimizing the KL divergence between the outputs generated from the original input and the masked input, in addition to the main loss function.
The effectiveness of the saliency approach has been demonstrated across various domains, including image analysis, language processing, and time series data \cite{tomar2022prequential,saadallah2022explainable}. 
Another saliency technique involves extracting a series of images from sliding windows within time series data and defining a learnable mask based on these series images and their perturbed counterparts. 
This approach, series saliency, acts as an adaptive data augmentation method for training deep models \cite{pan2021two}.
In exploring perturbed versions of data, Parvatharaju \textit{et al.} introduced a method called perturbation by prioritized replacement. 
This technique learns to emphasize the timesteps that contribute the most to the classifier's prediction, indicating their importance \cite{parvatharaju2021learning}.
Saadallah \textit{et al.} tackled the challenge of searching for an optimal network architecture by considering candidate models from various deep neural network architectures. 
They dynamically selected the most suitable architecture in real-time using concept drift detection in time series data. 
Saliency maps were utilized to compute the region of competence for each candidate network \cite{saadallah2021explainable}.

\section{Attention with Temporal Saliency}
\subsection{Problem Definition and Notation}

A time series is composed of N univariate time series where each $i=1 \dots N$, we have $y^i_t$ as a value of the univariate time series $i$ at time $t$. 
Given the look-back window $\tau$ , $x^i_{t}$ are exogenous inputs as associated co-variate values, e.g., day-of-the-week and hour-of-the-day. 
We can formulate the one-step-ahead prediction model as follows:

\begin{equation}
	\hat{y}_{i,t+1}=f(\ser{y}{i}{t}{\tau},\ser{x}{i}{t}{\tau})
\end{equation}
where $\ser{y}{i}{t}{\tau}=\{y^i_{t-\tau},...,y^i_{t}\}$ and $\ser{x}{i}{t}{\tau}=\{x^i_{t-\tau},...,x^i_{t}\}$.

As a common practice in transformer-based model, e.g., \cite{wang2020linformer} and \cite{xu2021autoformer}, the inputs $\ser{y}{i}{t}{\tau},\ser{x}{i}{t}{\tau}$ are encoded under a vector of hidden states $z$ to serve as the inputs for an attention block as the below step. 
The size $|z|$ is aligned with the number of input tokens for the transformer-based encoding block.

To prepare for the description of our architecture in the next section, we introduce two fundamental building blocks $\cD$ and $\cU$, with two following equations, e.g., downsampling and upsampling blocks, respectively. 
They are two parameterized sub-modules used in U-Net~\cite{ronneberger2015u} style architecture. 
They both have two parameters, namely, $\mathcal{H}$ and $d$, which are the hidden input states and the drop-out parameter.

\begin{equation}
	\cD(\mathcal{H},d)=\textrm{DropOut(MaxPool(ReLU(Conv1d}(\mathcal{H}))),d)
\end{equation}

\begin{equation}        
	\cU(\mathcal{H},d)=\\\textrm{DropOut(ConvT1d(ReLU(Conv1d}(\mathcal{H}))),d)
\end{equation}

\noindent
where ConvT1d is shorted for ConvTranspose1d.

\subsection{Architecture for Temporal Saliency Detection}
\begin{figure*}[t!]
	\centering
	\includegraphics[width=1.0\linewidth]{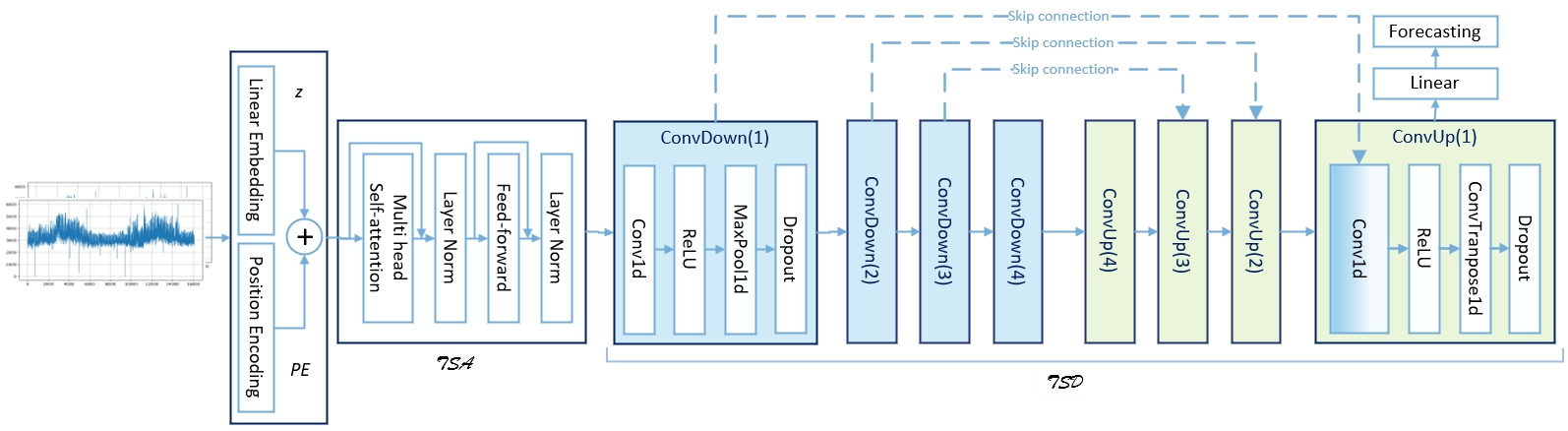} %
	\caption{The high-level design of the Temporal Saliency Detection (TSD) model.}
	\label{fig:funa-architecture}
\end{figure*}

This section will introduce our proposed learning architecture illustrated in Figure \ref{fig:funa-architecture}. 

The description will be followed from left to right according to the input flow. 
The critical novel aspect of this architecture is that the Temporal Saliency Detection ($\mathcal{TSD}$) block can work in tandem with the Temporal Self-Attention ($\mathcal{TSA}$) block. 
During the training process, the weights in both these blocks can be automatically adjusted to reveal temporal saliency maps in a similar fashion to semantic segmentation in computer vision. 
While our implementation is proven to outperform its competitors by a large margin, the saliency aspect will open the door for supporting the interpretability of our models as a natural next step for this work. 

\paragraph{Time Series Tokenization } To prepare the input for the below $\mathcal{TSA}$ block, our architecture will encode time series into a  Linear Embedding $z$ to make it compatible with token inputs of the attention module $\mathcal{TSA}$.
Next, we use convolution block \textrm{Conv1d} to encode $z$ from input time series $(\ser{y}{i}{t}{\tau},\ser{x}{i}{t}{\tau})$. 
Note the number of the tokens $|z|$ is a hyper-parameter for our architecture (see Ablation study in Section \ref{sec:abl}).

\begin{equation}
	z=\textrm{Conv1d}(\ser{y}{i}{t}{\tau},\ser{x}{i}{t}{\tau})
\end{equation}

\paragraph{Temporal Self-Attention ($\mathcal{TSA}$)} We use a multi-attention block to encode the correlation of temporal pattern, similar to \cite{wang2020linformer}. 
Here, we only use one multi-attention block to mitigate the memory problem of dot products similar to Informer and the like. 
The subsequent convolution blocks can be adjusted to avoid the quadratic memory consumption of multiple attention blocks of vanilla transformers. 
Our evaluation results and ablation study show that this block alone can work in concert with $\mathcal{TSD}$ to adjust the learning weights allowing temporal saliency patterns to emerge so that our trained models could outperform those of specification architectures in similar memory consumption. 
Also, as a common practice, we add the position encoding $\textrm{PE}$ \cite{vaswani2017attention} to $z$ to create $\mathcal{TSA}$ as follows.

\begin{equation}
	\mathcal{TSA}=\textrm{LN(FFN(LN(SelfAttention(}z+\textrm{PE))))}
\end{equation}

where $\textrm{LN(.), FFN(.)}$ and $\textrm{SelfAttention(.)}$ are LinearNorm, FeedForward and Self-Attention blocks, respectively.

\paragraph{Temporal Saliency Detection ($\mathcal{TSD}$)}
Inspired by saliency map generation using  U-Net in semantic segmentation, this block consists of two mirroring paths: contracting and expanding with $\cD$ and $\cU$ blocks, respectively. 
The below equations define such $\mathcal{L}$ block pairs. 
$\mathcal{L}$ is also considered as a hyper-parameter that can be empirically adjusted based on the data and the memory availability of the training infrastructure.
Figure~\ref{fig:funa-architecture} illustrates $\mathcal{L}=4$.

\begin{equation}        
	\textrm{ConvDown}^i=\cD\textrm{(ConvDown}^{i-1},d_i)
\end{equation}
where i=2,..,$\mathcal{L}$ and $\textrm{ConvDown}^1=\cD\mathcal(TSA,d_1)$.

\begin{equation}        
	\textrm{ConvUp}^i=\cU\textrm{(ConvDown}^i\oplus\textrm{ConvUp}^{i+1},d^i)
\end{equation}
where i=1,..,$\mathcal{L}$-1 and $\textrm{ConvUp}^{\mathcal{L}}=\cU(\textrm{ConvDown}^{\mathcal{L}},d^\mathcal{L})$. $\oplus$ is the concatenation operator.

Note that the skip connections are specified as the concatenations between $\textrm{ConvDown}^i$ and $\textrm{ConvUp}^{i+1}$. These skip connections are used to connect different patterns that emerged from different time scales. They also help avoid information loss due to the compression process that sparsification architectures suffer. Moreover, \cite{Wang0WZ22} indicated that such skip connections could be well integrated well with attention heads of $\mathcal{TSA}$. In this design, we can see the features for $z$ after the $\mathcal{TSA}$ might be at different scales or magnitudes. This can be due to some components of $z$ or later $\mathcal{TSA}$ having very sharp or very distributed attention weights  when summing over the features of the other components. Additionally, at the individual feature/vector entries level, concatenating across multiple attention heads—each of which might output values at different scales—can lead to the entries of the final vector having a wide range of values. Hence, these skip connections and the up-down sampling process work hand-in-hand to enable the temporal saliency patterns to emerge while canceling the noise. In the sequel, we have the definition of $\mathcal{TSD}$ block as follows.

\begin{equation}        
	\mathcal{TSD}=\textrm{LN(ConvUp}^1)
\end{equation}

Regarding $\mathcal{L}$ in our implementation for evaluated datasets, both contracting or expanding paths contain three or four repeated blocks, i.e., $\mathcal{L}=3$ or $\mathcal{L}=4$. 
Note that $\mathcal{L}$ can be seen as a counterpart of the $k$ parameters in 'top-k' components for sparsification architectures such as Informer and Autoformer. 
In our evaluation and ablation studies, $\mathcal{L}$ and associated parameters are more intuitive and easier to adjust to optimize the model performance empirically.

\paragraph{Forecasting} The forecasting operation 
involves one-step-ahead prediction $\hat{y}_{i,t+1}$ powered by $\mathcal{TSD}$ that can dynamically uses to compute a new hidden state $y^i_{t+1}$ for each element $i=1..N$ from the $\tau$ previous states  $\ser{y}{i}{t}{\tau}$ from $t$.

\section{Experiment}
\label{sec:experiment}

\subsection{Datasets}

To have a fair comparison with the current best approaches, we select the public data files from the Informer's Github page\footnote{\url{https://github.com/zhouhaoyi/Informer2020}}, including ETTh1, ETTh2, and ETTm1, and Autoformer's Github \footnote{\url{https://github.com/thuml/Autoformer}}, e.g., Exchange and ILI.
We evaluate all baselines and our model on a wide range of prediction horizons within \{24,36,48,60,96,168,288,336,672,720\}.

\textbf{ETT:} Electricity Transformer Temperature is a real-world dataset for electric power deployment.
The dataset is further converted into different granularity, e.g., ETTh1 and ETTh2 for 1-hour-level and ETTm1 for 15-minute-level.
Each data point consists of six predictors and one oil temperature target value.

\textbf{ILI:} Influenza-like illness dataset\footnote{\url{https://gis.cdc.gov/grasp/fluview/fluportaldashboard.html}} reports weekly recorded influenza patients from the Center for Disease Control and Prevention of the United States.
It measures the ratio of illness patients over the total number of patients in a week.
Each data point consists of six predictors and one target value.

\textbf{Exchange:} The dataset is a collection of daily exchange rates of different countries from 1990 to 2016 \cite{lai2018modeling}. 
Each data point consists of seven predictors and one target value.

\begin{figure*}[ht!]
	\centering
	\begin{subfigure}[t]{0.19\textwidth}
		\includegraphics[width=\textwidth]{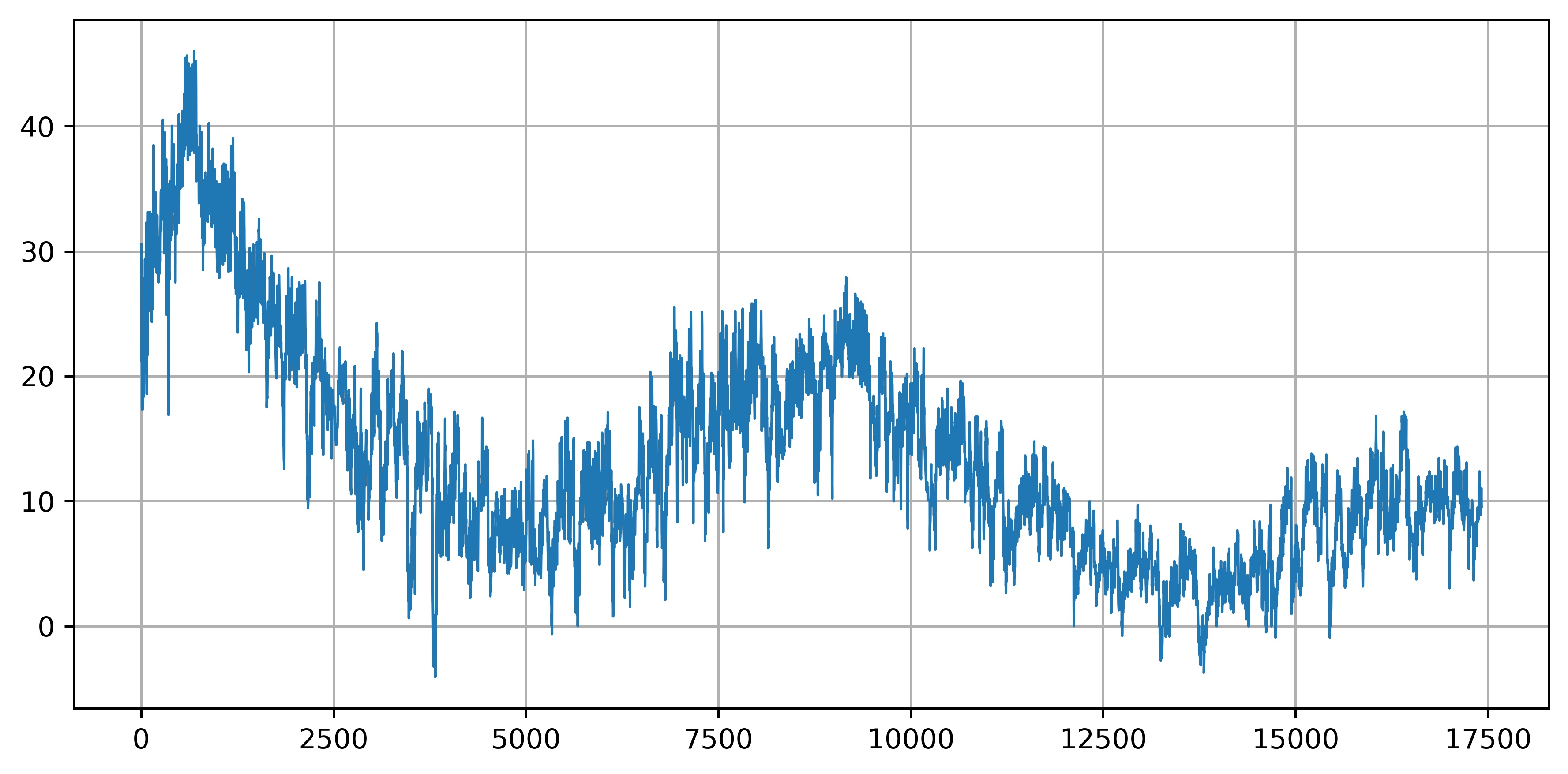}
		\caption{ETTh1}
		\label{subfig:ETTh1}
	\end{subfigure}
	\begin{subfigure}[t]{0.19\textwidth}
		\includegraphics[width=\textwidth]{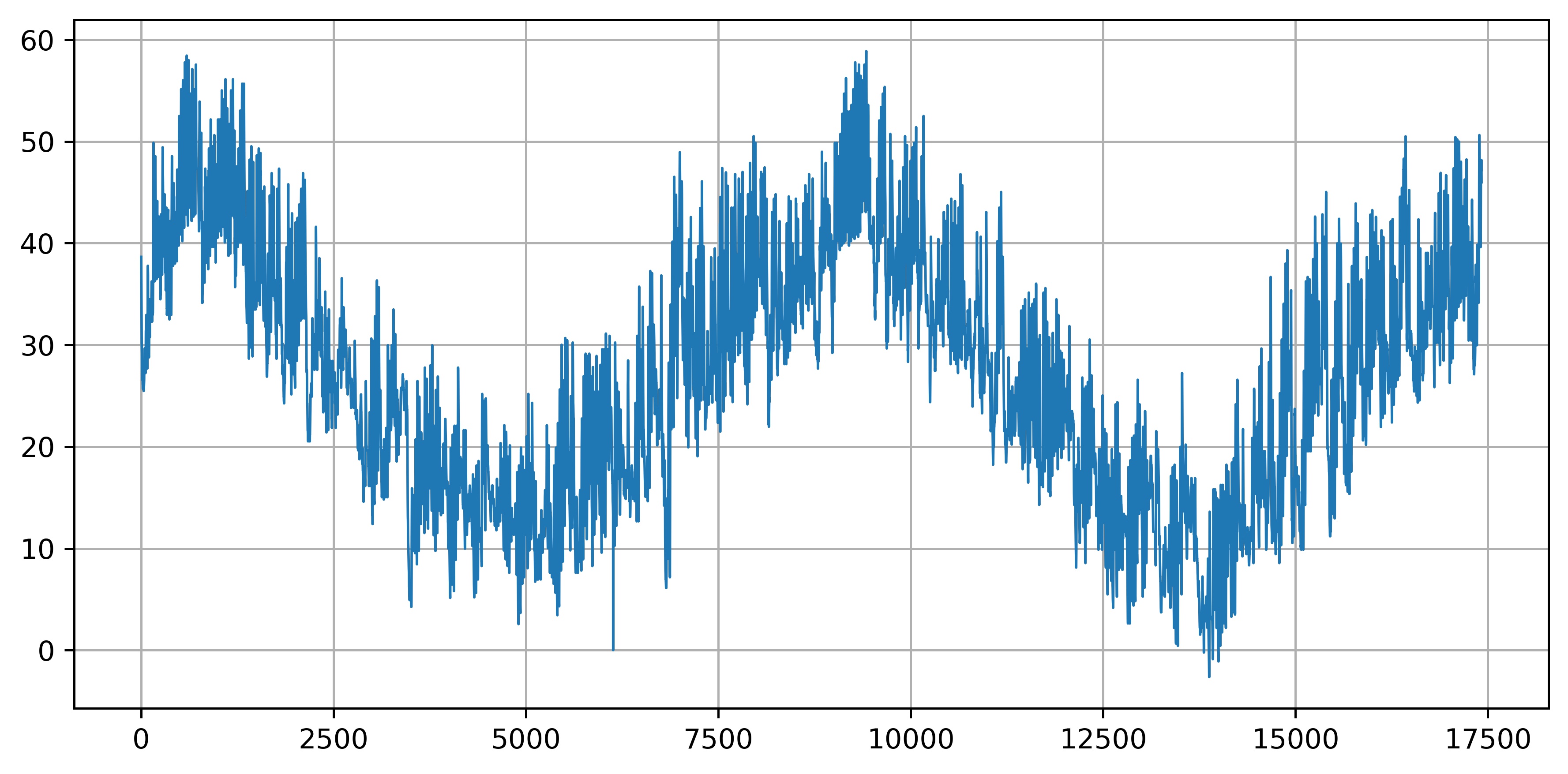}
		\caption{ETTh2}
		\label{subfig:ETTh2}
	\end{subfigure}
	\begin{subfigure}[t]{0.19\textwidth}
		\includegraphics[width=\textwidth]{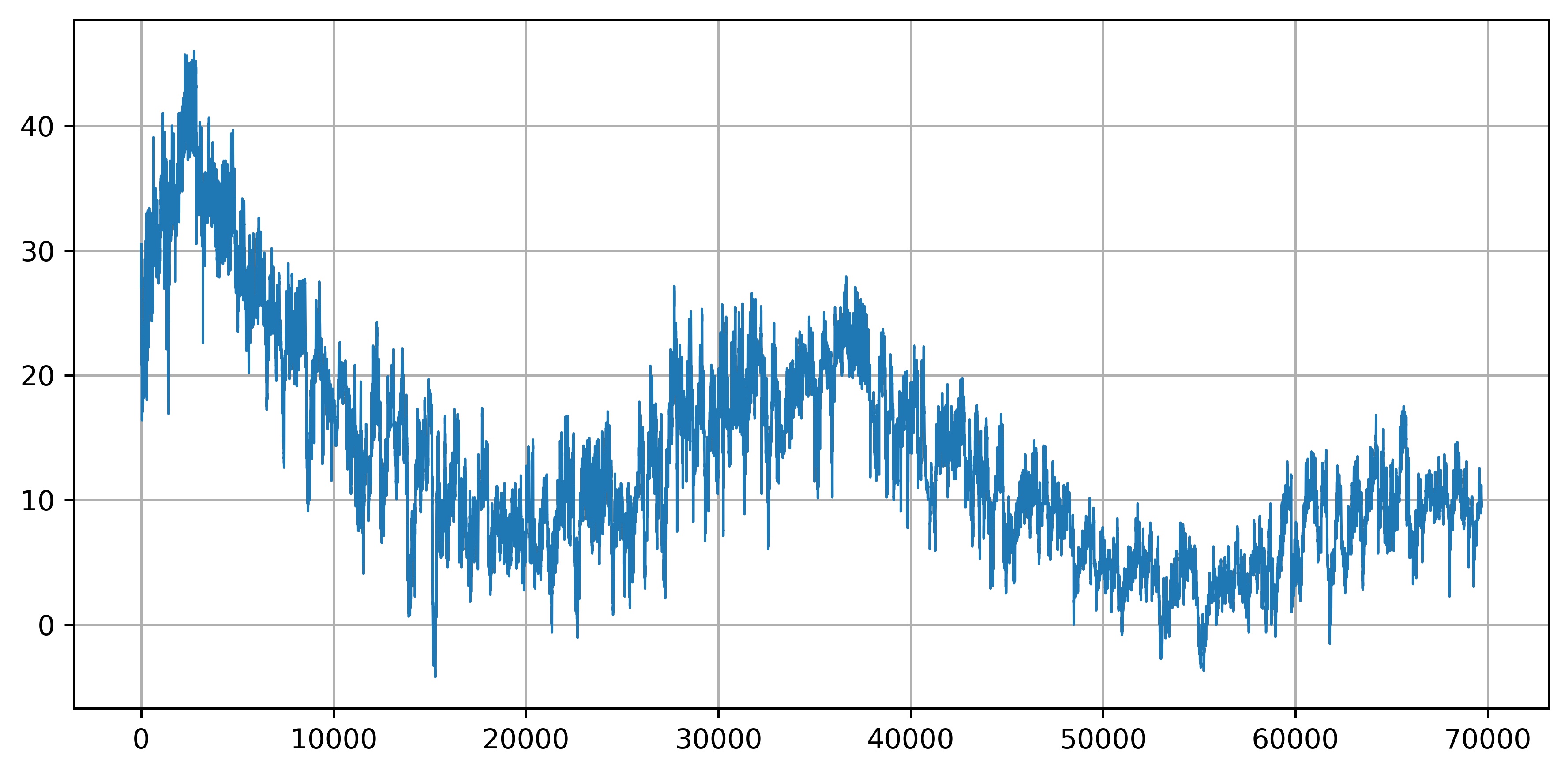}
		\caption{ETTm1}
		\label{subfig:ETTm1}
	\end{subfigure}
	\begin{subfigure}[t]{0.19\textwidth}
		\includegraphics[width=\textwidth]{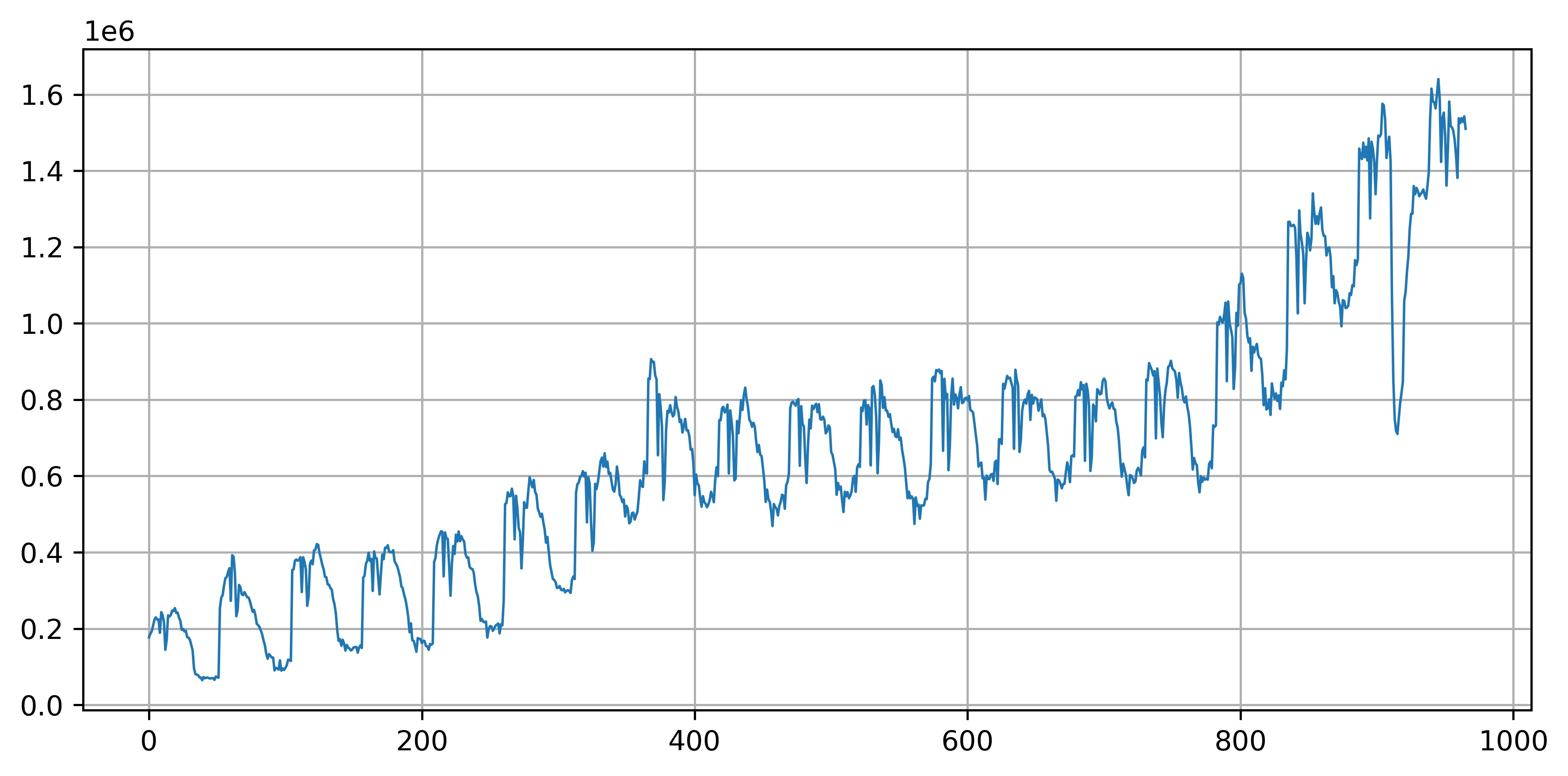}
		\caption{ILI}
		\label{subfig:ILI}
	\end{subfigure}
	\begin{subfigure}[t]{0.19\textwidth}
		\includegraphics[width=\textwidth]{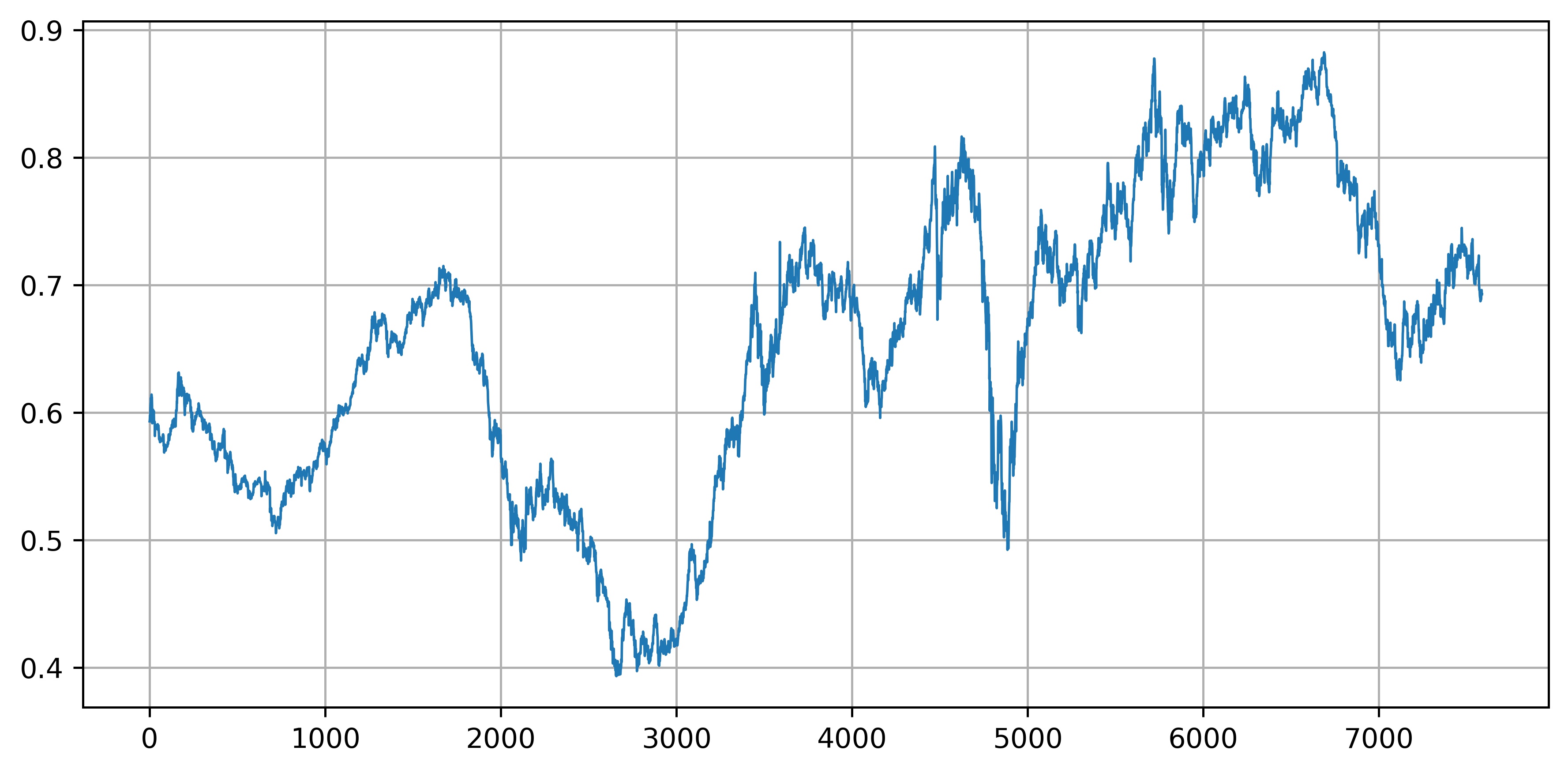}
		\caption{Exchange}
		\label{subfig:exchange}
	\end{subfigure}	
	\caption{A visualization of datasets' forecasting targets.}
	\label{fig:figures}
\end{figure*}

We follow the splitting protocol mentioned in the Autoformer paper \cite{xu2021autoformer} by the ratio of 7:1:2 for all datasets.
Figure \ref{fig:figures} visualizes the challenge of real-world finance Exchange, disease ILI, and energy consumption ETT datasets.
To be more precise, we choose time series that exhibit either a trend or are exceptionally challenging to predict.

\subsection{Experimental Setup}
\label{subsec:experimental-setup}
\textbf{Baselines}
We compared our TSD model with six baseline approaches.  
Regarding multivariate time series forecasting, we designed the experiments similar to the Autoformer model \cite{xu2021autoformer}, a standard benchmark for much following-up time series forecasting research.
More concretely, we compare TSD with Autoformer \cite{xu2021autoformer}, Informer \cite{zhou2021informer}, LogTrans \cite{li2019enhancing}, Reformer \cite{kitaev2020reformer}, LSTNet \cite{lai2018modeling}, and LSTM \cite{hochreiter1997long}.
We reused the experimental reports in the Autoformer paper after randomly double-checking by re-running several experiments not to repeat all implementations on all datasets.
Additionally, regarding univariate time series prediction, we designed the experiments similar to the Informer model \cite{zhou2021informer}.
Here, we compare our model with Informer, Reformer, LogTrans, DeepAR \cite{salinas2020deepar}, Yformer \cite{madhusudhanan2021yformer}, and Query Selector \cite{klimek2021long}.
The Git repository is available at \footnote{\url{https://github.com/duongtrung/time-series-temporal-saliency-patterns}}.

\textbf{Hyperparameter Optimization}
We conduct a grid search over the hyperparameters, and the ranges are given in the following.
We set the number of the encoder as one while mainly focusing on the Conv-down-up architecture design, which is the core power of the TSD model.
The hidden state sequence $z$ size is selected from $\{512,1024,1280,1536\}$ while the number of heads is $\{8,16,32\}$.
The number of up-sampling and down-sampling blocks is $\{3,4,5\}$.
The dropout values are searched from $\{0.05,0.1,0.2,0.3\}$.
We performed a grid search of the learning rate of $\{0.00001, 0.00002, 0.000001, 0.000005, 0.0000001\}$.
Furthermore, we apply a scheduling reduction of learning rate by a factor $\gamma$ of $\{0.97, 0.95, 0.87, 0.85\}$ with a step size of $\{2,5,8\}$.
The training epoch range is $\{15,20,25,30,35\}$.
The number of heads in self-attention is from $\{8,16\}$.
Regarding the optimizer, we select AdamW.

\textbf{Metrics and Implementation Details}
We trained the TSD model regarding optimizing the mean absolute error $ \textrm{MAE} = \frac{1}{n} \sum_{i=1}^{n} | \mathbf{y} - \hat{\mathbf{y}}| $ and mean squared error $ \textrm{MSE} = \frac{1}{n} \sum_{i=1}^{n} ( \mathbf{y} - \hat{\mathbf{y}})^2 $ losses on each prediction horizon.
We conducted most of Pytorch implementation on high-performance computing nodes equipped with a GeForce RTX 2080Ti 32GB.
We ran all experiments three times and reported the average results.

\subsection{Results and Analysis}
We select datasets under a wide range of horizon lengths from 24 to 720-time points to compare prediction capacity in the challenging multivariate scenario.
Regarding the multivariate time series forecasting, Table \ref{tab:result-af-mul} summarizes the experimental results of all models and benchmarks, while Figure \ref{fig:plot-comparison} visualizes a relative comparison in trends.
As for this experimental scenario, TSD has achieved remarkable performance in all benchmarks and all forecasting horizons.
Especially under the optimization for the MSE loss, TSD outperforms six baselines in all 23 different forecasting lengths.
Autoformer achieves better results in only 4 MAE cases out of 46 cases in both MSE and MAE.
However, these better results fall into short forecasting horizons, e.g., (48, 168) and (96, 192) in ETTh2 and Exchange, respectively. 
The trends of ETTh1, ETTh2, ILI, and ETTm1 in Figure \ref{fig:plot-comparison} have proven the consistent prediction power of TSD in far horizons.
Especially, under the MSE loss, compared to the previous best state-of-the-art performance, TSD gives average \textbf{65\%} reduction in ETTh1, \textbf{29\%} in ETTh2, \textbf{65\%} in ETTm1, \textbf{18\%} in Exchange, and \textbf{56\%} in ILI.
A similar reduction is observed regarding the MAE loss.
The error reduction harmonizes with trends and nature of the data as shown in Figure \ref{fig:plot-comparison}.
ETTh1 and ETTm1 share a general downward trend, and the data part corresponding to the test set has a small fluctuation amplitude.
Contrary to ETTh1 and ETTm1, the ILI dataset tends to increase, and the corresponding test set data has a larger fluctuation amplitude.
Due to the unforeseen interaction between countless economic phenomena, the Exchange dataset is comparatively challenging.  
Therefore, a reduction of \textbf{18\%} is achieved.
Overall, 46 reported results in 10 different forecasting horizons, TSD yields a \textbf{31\%} reduction in both MSE and MAE losses.
It proves TSD's superiority against several current best models for many complex real-world multivariate forecasting applications, including early disease warning, long-term financial planning, and resource consumption arrangement.

As for the univariate scenario, i.e $N=1$, TSD has also achieved consistent performance in all benchmarks and all forecasting horizons.
TSD yields the best scores in 28 out of 30 experimental cases within five different horizons.
Note that TSD has noticeably outperformed DeepAR, the back-bone model for Amazon Forecast Service.
The Query Selector model is a robust approach outperforming Informer in all three datasets in both losses.
However, compared to our model, the Query Selector is only better in two cases, e.g., ETTm1 at 672 horizons for MSE and MAE.
The relative error reduction in both MSE and MAE losses is \textbf{46\%}.

\begin{table*}[ht!]
	\centering
	\caption{\textbf{Multivariate} time series forecasting results. A lower MSE or MAE indicates better forecasting. The best scores are in \textbf{bold}.}
	\label{tab:result-af-mul}
	\resizebox{0.99\textwidth}{!}{%
		\begin{tabular}{|ll|cc|cc|cc|cc|cc|cc|cc|}
			\hline
			\multicolumn{2}{|c|}{Model}                           & \multicolumn{2}{c|}{TSD}                             & \multicolumn{2}{c|}{Autoformer}             & \multicolumn{2}{c|}{Informer}      & \multicolumn{2}{c|}{LogTrans}      & \multicolumn{2}{c|}{Reformer}      & \multicolumn{2}{c|}{LSTNet}        & \multicolumn{2}{c|}{LSTM}          \\ \hline
			\multicolumn{2}{|c|}{Data \& Metric}                  & \multicolumn{1}{c|}{MSE}            & MAE            & \multicolumn{1}{c|}{MSE}   & MAE            & \multicolumn{1}{c|}{MSE}   & MAE   & \multicolumn{1}{c|}{MSE}   & MAE   & \multicolumn{1}{c|}{MSE}   & MAE   & \multicolumn{1}{c|}{MSE}   & MAE   & \multicolumn{1}{c|}{MSE}   & MAE   \\ \hline
			\multicolumn{1}{|l|}{\multirow{5}{*}{ETTh1}}    & 24  & \multicolumn{1}{c|}{\textbf{0.115}} & \textbf{0.285} & \multicolumn{1}{c|}{0.384} & 0.425          & \multicolumn{1}{c|}{0.577} & 0.549 & \multicolumn{1}{c|}{0.686} & 0.604 & \multicolumn{1}{c|}{0.991} & 0.754 & \multicolumn{1}{c|}{1.293} & 0.901 & \multicolumn{1}{c|}{0.650} & 0.624 \\ \cline{2-16} 
			\multicolumn{1}{|l|}{}                          & 48  & \multicolumn{1}{c|}{\textbf{0.119}} & \textbf{0.279} & \multicolumn{1}{c|}{0.392} & 0.419          & \multicolumn{1}{c|}{0.685} & 0.625 & \multicolumn{1}{c|}{0.766} & 0.757 & \multicolumn{1}{c|}{1.313} & 0.906 & \multicolumn{1}{c|}{1.456} & 0.960 & \multicolumn{1}{c|}{0.702} & 0.675 \\ \cline{2-16} 
			\multicolumn{1}{|l|}{}                          & 168 & \multicolumn{1}{c|}{\textbf{0.191}} & \textbf{0.313} & \multicolumn{1}{c|}{0.490} & 0.481          & \multicolumn{1}{c|}{0.931} & 0.752 & \multicolumn{1}{c|}{1.002} & 0.846 & \multicolumn{1}{c|}{1.824} & 1.138 & \multicolumn{1}{c|}{1.997} & 1.214 & \multicolumn{1}{c|}{1.212} & 0.867 \\ \cline{2-16} 
			\multicolumn{1}{|l|}{}                          & 336 & \multicolumn{1}{c|}{\textbf{0.141}} & \textbf{0.285} & \multicolumn{1}{c|}{0.505} & 0.484          & \multicolumn{1}{c|}{1.128} & 0.873 & \multicolumn{1}{c|}{1.362} & 0.952 & \multicolumn{1}{c|}{2.117} & 1.280 & \multicolumn{1}{c|}{2.655} & 1.369 & \multicolumn{1}{c|}{1.424} & 0.994 \\ \cline{2-16} 
			\multicolumn{1}{|l|}{}                          & 720 & \multicolumn{1}{c|}{\textbf{0.233}} & \textbf{0.416} & \multicolumn{1}{c|}{0.498} & 0.500          & \multicolumn{1}{c|}{1.215} & 0.896 & \multicolumn{1}{c|}{1.397} & 1.291 & \multicolumn{1}{c|}{2.415} & 1.520 & \multicolumn{1}{c|}{2.143} & 1.380 & \multicolumn{1}{c|}{1.960} & 1.322 \\ \hline
			\multicolumn{1}{|l|}{\multirow{5}{*}{ETTh2}}    & 24  & \multicolumn{1}{c|}{\textbf{0.189}} & \textbf{0.270} & \multicolumn{1}{c|}{0.261} & 0.341          & \multicolumn{1}{c|}{0.720} & 0.665 & \multicolumn{1}{c|}{0.828} & 0.750 & \multicolumn{1}{c|}{1.531} & 1.613 & \multicolumn{1}{c|}{2.742} & 1.457 & \multicolumn{1}{c|}{1.143} & 0.813 \\ \cline{2-16} 
			\multicolumn{1}{|l|}{}                          & 48  & \multicolumn{1}{c|}{\textbf{0.242}} & 0.538          & \multicolumn{1}{c|}{0.312} & \textbf{0.373} & \multicolumn{1}{c|}{1.457} & 1.001 & \multicolumn{1}{c|}{1.806} & 1.034 & \multicolumn{1}{c|}{1.871} & 1.735 & \multicolumn{1}{c|}{3.567} & 1.687 & \multicolumn{1}{c|}{1.671} & 1.221 \\ \cline{2-16} 
			\multicolumn{1}{|l|}{}                          & 168 & \multicolumn{1}{c|}{\textbf{0.320}} & 0.493          & \multicolumn{1}{c|}{0.457} & \textbf{0.455} & \multicolumn{1}{c|}{3.489} & 1.515 & \multicolumn{1}{c|}{4.070} & 1.681 & \multicolumn{1}{c|}{4.660} & 1.846 & \multicolumn{1}{c|}{3.242} & 2.513 & \multicolumn{1}{c|}{4.117} & 1.674 \\ \cline{2-16} 
			\multicolumn{1}{|l|}{}                          & 336 & \multicolumn{1}{c|}{\textbf{0.309}} & \textbf{0.462} & \multicolumn{1}{c|}{0.471} & 0.475          & \multicolumn{1}{c|}{2.723} & 1.340 & \multicolumn{1}{c|}{3.875} & 1.763 & \multicolumn{1}{c|}{4.028} & 1.688 & \multicolumn{1}{c|}{2.544} & 2.591 & \multicolumn{1}{c|}{3.434} & 1.549 \\ \cline{2-16} 
			\multicolumn{1}{|l|}{}                          & 720 & \multicolumn{1}{c|}{\textbf{0.314}} & \textbf{0.472} & \multicolumn{1}{c|}{0.474} & 0.484          & \multicolumn{1}{c|}{3.467} & 1.473 & \multicolumn{1}{c|}{3.913} & 1.552 & \multicolumn{1}{c|}{5.381} & 2.015 & \multicolumn{1}{c|}{4.625} & 3.709 & \multicolumn{1}{c|}{3.963} & 1.788 \\ \hline
			\multicolumn{1}{|l|}{\multirow{5}{*}{ETTm1}}    & 24  & \multicolumn{1}{c|}{\textbf{0.145}} & \textbf{0.306} & \multicolumn{1}{c|}{0.383} & 0.403          & \multicolumn{1}{c|}{0.323} & 0.369 & \multicolumn{1}{c|}{0.419} & 0.412 & \multicolumn{1}{c|}{0.724} & 0.607 & \multicolumn{1}{c|}{1.968} & 1.170 & \multicolumn{1}{c|}{0.621} & 0.629 \\ \cline{2-16} 
			\multicolumn{1}{|l|}{}                          & 48  & \multicolumn{1}{c|}{\textbf{0.143}} & \textbf{0.293} & \multicolumn{1}{c|}{0.454} & 0.453          & \multicolumn{1}{c|}{0.494} & 0.503 & \multicolumn{1}{c|}{0.507} & 0.583 & \multicolumn{1}{c|}{1.098} & 0.777 & \multicolumn{1}{c|}{1.999} & 1.215 & \multicolumn{1}{c|}{1.392} & 0.939 \\ \cline{2-16} 
			\multicolumn{1}{|l|}{}                          & 96  & \multicolumn{1}{c|}{\textbf{0.268}} & \textbf{0.390} & \multicolumn{1}{c|}{0.481} & 0.463          & \multicolumn{1}{c|}{0.678} & 0.614 & \multicolumn{1}{c|}{0.768} & 0.792 & \multicolumn{1}{c|}{1.433} & 0.945 & \multicolumn{1}{c|}{2.762} & 1.542 & \multicolumn{1}{c|}{1.339} & 0.913 \\ \cline{2-16} 
			\multicolumn{1}{|l|}{}                          & 288 & \multicolumn{1}{c|}{\textbf{0.157}} & \textbf{0.316} & \multicolumn{1}{c|}{0.634} & 0.528          & \multicolumn{1}{c|}{1.056} & 0.786 & \multicolumn{1}{c|}{1.462} & 1.320 & \multicolumn{1}{c|}{1.820} & 1.094 & \multicolumn{1}{c|}{1.257} & 2.076 & \multicolumn{1}{c|}{1.740} & 1.124 \\ \cline{2-16} 
			\multicolumn{1}{|l|}{}                          & 672 & \multicolumn{1}{c|}{\textbf{0.149}} & \textbf{0.313} & \multicolumn{1}{c|}{0.606} & 0.542          & \multicolumn{1}{c|}{1.192} & 0.926 & \multicolumn{1}{c|}{1.669} & 1.461 & \multicolumn{1}{c|}{2.187} & 1.232 & \multicolumn{1}{c|}{1.917} & 2.941 & \multicolumn{1}{c|}{2.736} & 1.555 \\ \hline
			\multicolumn{1}{|l|}{\multirow{4}{*}{Exchange}} & 96  & \multicolumn{1}{c|}{\textbf{0.184}} & 0.369          & \multicolumn{1}{c|}{0.197} & \textbf{0.323} & \multicolumn{1}{c|}{0.847} & 0.752 & \multicolumn{1}{c|}{0.968} & 0.812 & \multicolumn{1}{c|}{1.065} & 0.829 & \multicolumn{1}{c|}{1.551} & 1.058 & \multicolumn{1}{c|}{1.453} & 1.049 \\ \cline{2-16} 
			\multicolumn{1}{|l|}{}                          & 192 & \multicolumn{1}{c|}{\textbf{0.262}} & 0.445          & \multicolumn{1}{c|}{0.300} & \textbf{0.369} & \multicolumn{1}{c|}{1.204} & 0.895 & \multicolumn{1}{c|}{1.040} & 0.851 & \multicolumn{1}{c|}{1.188} & 0.906 & \multicolumn{1}{c|}{1.477} & 1.028 & \multicolumn{1}{c|}{1.846} & 1.179 \\ \cline{2-16} 
			\multicolumn{1}{|l|}{}                          & 336 & \multicolumn{1}{c|}{\textbf{0.293}} & \textbf{0.422} & \multicolumn{1}{c|}{0.509} & 0.524          & \multicolumn{1}{c|}{1.672} & 1.036 & \multicolumn{1}{c|}{1.659} & 1.081 & \multicolumn{1}{c|}{1.357} & 0.976 & \multicolumn{1}{c|}{1.507} & 1.031 & \multicolumn{1}{c|}{2.136} & 1.231 \\ \cline{2-16} 
			\multicolumn{1}{|l|}{}                          & 720 & \multicolumn{1}{c|}{\textbf{1.307}} & \textbf{0.758} & \multicolumn{1}{c|}{1.447} & 0.941          & \multicolumn{1}{c|}{2.478} & 1.310 & \multicolumn{1}{c|}{1.941} & 1.127 & \multicolumn{1}{c|}{1.510} & 1.016 & \multicolumn{1}{c|}{2.285} & 1.243 & \multicolumn{1}{c|}{2.984} & 1.427 \\ \hline
			\multicolumn{1}{|l|}{\multirow{4}{*}{ILI}}      & 24  & \multicolumn{1}{c|}{\textbf{1.514}} & \textbf{1.103} & \multicolumn{1}{c|}{3.483} & 1.287          & \multicolumn{1}{c|}{5.764} & 1.677 & \multicolumn{1}{c|}{4.480} & 1.444 & \multicolumn{1}{c|}{4.400} & 1.382 & \multicolumn{1}{c|}{6.026} & 1.770 & \multicolumn{1}{c|}{5.914} & 1.734 \\ \cline{2-16} 
			\multicolumn{1}{|l|}{}                          & 36  & \multicolumn{1}{c|}{\textbf{1.449}} & \textbf{1.086} & \multicolumn{1}{c|}{3.103} & 1.148          & \multicolumn{1}{c|}{4.755} & 1.467 & \multicolumn{1}{c|}{4.799} & 1.467 & \multicolumn{1}{c|}{4.783} & 1.448 & \multicolumn{1}{c|}{5.340} & 1.668 & \multicolumn{1}{c|}{6.631} & 1.845 \\ \cline{2-16} 
			\multicolumn{1}{|l|}{}                          & 48  & \multicolumn{1}{c|}{\textbf{1.186}} & \textbf{0.971} & \multicolumn{1}{c|}{2.669} & 1.085          & \multicolumn{1}{c|}{4.763} & 1.469 & \multicolumn{1}{c|}{4.800} & 1.468 & \multicolumn{1}{c|}{4.832} & 1.465 & \multicolumn{1}{c|}{6.080} & 1.787 & \multicolumn{1}{c|}{6.736} & 1.857 \\ \cline{2-16} 
			\multicolumn{1}{|l|}{}                          & 60  & \multicolumn{1}{c|}{\textbf{1.140}} & \textbf{0.946} & \multicolumn{1}{c|}{2.770} & 1.125          & \multicolumn{1}{c|}{5.264} & 1.564 & \multicolumn{1}{c|}{5.278} & 1.560 & \multicolumn{1}{c|}{4.882} & 1.483 & \multicolumn{1}{c|}{5.548} & 1.720 & \multicolumn{1}{c|}{6.870} & 1.879 \\ \hline
		\end{tabular}%
	}
\end{table*}

\subsection{Discussions}

When applying prediction principles to sequential data, e.g., natural language or time series, contextual information weighs a lot, primarily when long-range dependencies exist. 
In this context, the issues of gradient vanishing and explosion, model size, and dependencies depend on the length of the sequential data. 
Transformer's self-attention approach successfully addressed those mentioned issues by designing a novel encoder-decoder architecture \cite{vaswani2017attention}.
Developed upon that architecture, the Probsparse self-attention was introduced to overcome the memory bottleneck of the transformer while acceptably handling extremely long input sequences. 
Table \ref{tab:simplification} presents the evolutional simplification of general encoder-decoder pairs throughout the research. 
Employing model design, the complexity of an encoder-decoder architecture is significantly reduced.
For instance, the number of encoders and decoders in the Informer model was reduced from 6 to 4 and 6 to 2, respectively. 
However, the long-term forecasting problem of time series remains challenging, although various self-attention mechanisms were adopted.
In far-horizon forecasting, a model, instead of attending to several single points, treat sub-series level and aggregates dependencies discovery and representation.
Consequently, the auto-correlation mechanism was developed \cite{xu2021autoformer}, which yielded a 38\% relative improvement on compared models.
The number of encoders and decoders was noticeably reduced from 4 to 2 and 2 to 1, respectively.
One point to note is that the authors of those mentioned models do not provide any ablation study on why they chose the number of encoders and decoders. However, the core idea is to balance forecasting accuracy and computation efficiency.
As discussed in Section \ref{sec:introduction}, our hypothesis is to automatically encode the correct temporal pattern to the suitable self-attention heads and to learn saliency patterns emerging from the sequential data.
Hence, one encoder is enough to output a self-attention representation.
Unlike the existing methods, we completely replace a general decoder with a U-sharp architecture, effectively addressing the image segmentation task.
Generally speaking, we also want to \textit{segment} time series in automatic processing and discovery.

\begin{table*}[ht!]
	\centering
	\caption{\textbf{Univariate} time series forecasting results. A lower MSE or MAE indicates better forecasting. The best scores are in bold. }
	\label{tab:uni}
	\resizebox{0.99\textwidth}{!}{%
		\begin{tabular}{|ll|cc|cc|cc|cc|cc|cc|cc|}
			\hline
			\multicolumn{2}{|c|}{Models}                       & \multicolumn{2}{c|}{TSD}                             & \multicolumn{2}{c|}{Informer}      & \multicolumn{2}{c|}{Reformer}      & \multicolumn{2}{c|}{LogTrans}      & \multicolumn{2}{c|}{DeepAR}        & \multicolumn{2}{c|}{Yformer}                & \multicolumn{2}{c|}{\begin{tabular}[c]{@{}c@{}}Query\\ Selector\end{tabular}} \\ \hline
			\multicolumn{2}{|c|}{Dataset \& Metric}            & \multicolumn{1}{c|}{MSE}            & MAE            & \multicolumn{1}{c|}{MSE}   & MAE   & \multicolumn{1}{c|}{MSE}   & MAE   & \multicolumn{1}{c|}{MSE}   & MAE   & \multicolumn{1}{c|}{MSE}   & MAE   & \multicolumn{1}{c|}{MSE}            & MAE   & \multicolumn{1}{c|}{MSE}                         & MAE                        \\ \hline
			\multicolumn{1}{|l|}{\multirow{5}{*}{ETTh1}} & 24  & \multicolumn{1}{c|}{\textbf{0.018}} & \textbf{0.102} & \multicolumn{1}{c|}{0.098} & 0.247 & \multicolumn{1}{c|}{0.222} & 0.389 & \multicolumn{1}{c|}{0.103} & 0.259 & \multicolumn{1}{c|}{0.107} & 0.280 & \multicolumn{1}{c|}{0.082}          & 0.230 & \multicolumn{1}{c|}{\textit{0.043}}              & 0.161                      \\ \cline{2-16} 
			\multicolumn{1}{|l|}{}                       & 48  & \multicolumn{1}{c|}{\textbf{0.043}} & \textbf{0.166} & \multicolumn{1}{c|}{0.158} & 0.319 & \multicolumn{1}{c|}{0.284} & 0.445 & \multicolumn{1}{c|}{0.167} & 0.328 & \multicolumn{1}{c|}{0.162} & 0.327 & \multicolumn{1}{c|}{0.139}          & 0.308 & \multicolumn{1}{c|}{\textit{0.072}}              & 0.211                      \\ \cline{2-16} 
			\multicolumn{1}{|l|}{}                       & 168 & \multicolumn{1}{c|}{\textbf{0.082}} & \textbf{0.225} & \multicolumn{1}{c|}{0.183} & 0.346 & \multicolumn{1}{c|}{1.522} & 1.191 & \multicolumn{1}{c|}{0.207} & 0.375 & \multicolumn{1}{c|}{0.239} & 0.422 & \multicolumn{1}{c|}{0.111}          & 0.268 & \multicolumn{1}{c|}{\textit{0.093}}              & 0.237                      \\ \cline{2-16} 
			\multicolumn{1}{|l|}{}                       & 336 & \multicolumn{1}{c|}{\textbf{0.094}} & \textbf{0.237} & \multicolumn{1}{c|}{0.222} & 0.387 & \multicolumn{1}{c|}{1.860} & 1.124 & \multicolumn{1}{c|}{0.230} & 0.398 & \multicolumn{1}{c|}{0.445} & 0.552 & \multicolumn{1}{c|}{0.195}          & 0.365 & \multicolumn{1}{c|}{\textit{0.126}}              & 0.284                      \\ \cline{2-16} 
			\multicolumn{1}{|l|}{}                       & 720 & \multicolumn{1}{c|}{\textbf{0.129}} & \textbf{0.291} & \multicolumn{1}{c|}{0.269} & 0.435 & \multicolumn{1}{c|}{2.112} & 1.436 & \multicolumn{1}{c|}{0.273} & 0.463 & \multicolumn{1}{c|}{0.658} & 0.707 & \multicolumn{1}{c|}{0.226}          & 0.394 & \multicolumn{1}{c|}{\textit{0.213}}              & 0.373                      \\ \hline
			\multicolumn{1}{|l|}{\multirow{5}{*}{ETTm1}} & 24  & \multicolumn{1}{c|}{\textbf{0.011}} & \textbf{0.082} & \multicolumn{1}{c|}{0.030} & 0.137 & \multicolumn{1}{c|}{0.095} & 0.228 & \multicolumn{1}{c|}{0.065} & 0.202 & \multicolumn{1}{c|}{0.091} & 0.243 & \multicolumn{1}{c|}{0.024}          & 0.118 & \multicolumn{1}{c|}{\textit{0.013}}              & 0.087                      \\ \cline{2-16} 
			\multicolumn{1}{|l|}{}                       & 48  & \multicolumn{1}{c|}{\textbf{0.019}} & \textbf{0.110} & \multicolumn{1}{c|}{0.069} & 0.203 & \multicolumn{1}{c|}{0.249} & 0.390 & \multicolumn{1}{c|}{0.078} & 0.220 & \multicolumn{1}{c|}{0.219} & 0.362 & \multicolumn{1}{c|}{0.048}          & 0.173 & \multicolumn{1}{c|}{\textit{0.034}}              & 0.140                      \\ \cline{2-16} 
			\multicolumn{1}{|l|}{}                       & 96  & \multicolumn{1}{c|}{\textbf{0.038}} & \textbf{0.161} & \multicolumn{1}{c|}{0.194} & 0.372 & \multicolumn{1}{c|}{0.920} & 0.767 & \multicolumn{1}{c|}{0.199} & 0.386 & \multicolumn{1}{c|}{0.364} & 0.496 & \multicolumn{1}{c|}{0.143}          & 0.311 & \multicolumn{1}{c|}{\textit{0.070}}              & 0.210                      \\ \cline{2-16} 
			\multicolumn{1}{|l|}{}                       & 288 & \multicolumn{1}{c|}{\textbf{0.057}} & \textbf{0.199} & \multicolumn{1}{c|}{0.401} & 0.554 & \multicolumn{1}{c|}{1.108} & 1.245 & \multicolumn{1}{c|}{0.411} & 0.572 & \multicolumn{1}{c|}{0.948} & 0.795 & \multicolumn{1}{c|}{\textit{0.150}} & 0.316 & \multicolumn{1}{c|}{0.154}                       & 0.324                      \\ \cline{2-16} 
			\multicolumn{1}{|l|}{}                       & 672 & \multicolumn{1}{c|}{0.341}          & 1.052          & \multicolumn{1}{c|}{0.512} & 0.644 & \multicolumn{1}{c|}{1.793} & 1.528 & \multicolumn{1}{c|}{0.598} & 0.702 & \multicolumn{1}{c|}{2.437} & 1.352 & \multicolumn{1}{c|}{0.305}          & 0.476 & \multicolumn{1}{c|}{\textbf{0.173}}              & \textbf{0.342}             \\ \hline
			\multicolumn{1}{|l|}{\multirow{5}{*}{ETTh2}} & 24  & \multicolumn{1}{c|}{\textbf{0.075}} & \textbf{0.210} & \multicolumn{1}{c|}{0.093} & 0.240 & \multicolumn{1}{c|}{0.263} & 0.437 & \multicolumn{1}{c|}{0.102} & 0.255 & \multicolumn{1}{c|}{0.098} & 0.263 & \multicolumn{1}{c|}{\textit{0.082}} & 0.221 & \multicolumn{1}{c|}{0.084}                       & 0.223                      \\ \cline{2-16} 
			\multicolumn{1}{|l|}{}                       & 48  & \multicolumn{1}{c|}{\textbf{0.073}} & \textbf{0.213} & \multicolumn{1}{c|}{0.155} & 0.314 & \multicolumn{1}{c|}{0.458} & 0.545 & \multicolumn{1}{c|}{0.169} & 0.348 & \multicolumn{1}{c|}{0.163} & 0.341 & \multicolumn{1}{c|}{0.139}          & 0.334 & \multicolumn{1}{c|}{\textit{0.111}}              & 0.262                      \\ \cline{2-16} 
			\multicolumn{1}{|l|}{}                       & 168 & \multicolumn{1}{c|}{\textbf{0.110}} & \textbf{0.270} & \multicolumn{1}{c|}{0.232} & 0.389 & \multicolumn{1}{c|}{1.029} & 0.879 & \multicolumn{1}{c|}{0.246} & 0.422 & \multicolumn{1}{c|}{0.255} & 0.414 & \multicolumn{1}{c|}{\textit{0.111}} & 0.337 & \multicolumn{1}{c|}{0.175}                       & 0.332                      \\ \cline{2-16} 
			\multicolumn{1}{|l|}{}                       & 336 & \multicolumn{1}{c|}{\textbf{0.121}} & \textbf{0.273} & \multicolumn{1}{c|}{0.263} & 0.417 & \multicolumn{1}{c|}{1.668} & 1.228 & \multicolumn{1}{c|}{0.267} & 0.437 & \multicolumn{1}{c|}{0.604} & 0.607 & \multicolumn{1}{c|}{\textit{0.195}} & 0.391 & \multicolumn{1}{c|}{0.208}                       & 0.371                      \\ \cline{2-16} 
			\multicolumn{1}{|l|}{}                       & 720 & \multicolumn{1}{c|}{\textbf{0.123}} & \textbf{0.273} & \multicolumn{1}{c|}{0.277} & 0.431 & \multicolumn{1}{c|}{2.030} & 1.721 & \multicolumn{1}{c|}{0.303} & 0.493 & \multicolumn{1}{c|}{0.429} & 0.580 & \multicolumn{1}{c|}{\textit{0.226}} & 0.382 & \multicolumn{1}{c|}{0.258}                       & 0.413                      \\ \hline
		\end{tabular}%
	}
\end{table*}


\section{Ablation Study}
\label{sec:abl}


\subsection{Pooling Selection}
TSD comprises $u$-pair blocks of down-sampling and up-sampling that play a central role in current architecture.
At each block $l$, we propose to use a pooling layer with a kernel size of $k_l = 3$ to help the layer focus a specific scale if its input.
It helps reduce the input's width, release memory usage, reduce learnable parameters, alleviate the effects of overfitting, and limit the computation. 
We carefully explore max and average pooling operations and the model's stability under various far-horizon.
Table \ref{tab:pooling-config} presents the empirical evaluation of pooling configurations.
This ablation shows that max pooling is more stable in long-term forecasting, e.g., from a horizon of 60 and beyond.
We have randomly tested on other datasets and observed a similar trend, more details can be founded in supplementary materials.

\begin{table}[ht!]
	\caption{Empirical evaluation of eight different horizons with two pooling settings. The best scores are in \textbf{bold}. }
	\label{tab:pooling-config}
	\centering
		\begin{tabular}{|l|l|cc|cc|}
			\hline
			\multicolumn{1}{|c|}{\multirow{2}{*}{Data}} & \multicolumn{1}{c|}{\multirow{2}{*}{Horizon}} & \multicolumn{2}{c|}{MaxPool}                         & \multicolumn{2}{c|}{AveragePool}                     \\ \cline{3-6} 
			\multicolumn{1}{|c|}{}                      & \multicolumn{1}{c|}{}                         & \multicolumn{1}{c|}{MSE}            & MAE            & \multicolumn{1}{c|}{MSE}            & MAE            \\ \hline
			\multirow{4}{*}{ILI}                        & 24                                            & \multicolumn{1}{c|}{\textbf{1.514}} & \textbf{1.103} & \multicolumn{1}{c|}{1.577}          & 1.108          \\ \cline{2-6} 
			& 36                                            & \multicolumn{1}{c|}{1.449}          & 1.086          & \multicolumn{1}{c|}{\textbf{1.429}} & \textbf{1.042} \\ \cline{2-6} 
			& 48                                            & \multicolumn{1}{c|}{\textbf{1.186}} & 0.971          & \multicolumn{1}{c|}{1.239}          & \textbf{0.962} \\ \cline{2-6} 
			& 60                                            & \multicolumn{1}{c|}{\textbf{1.140}} & \textbf{0.946} & \multicolumn{1}{c|}{1.146}          & 0.947          \\ \hline
			\multirow{4}{*}{Exchange}                   & 96                                            & \multicolumn{1}{c|}{\textbf{0.184}} & \textbf{0.369} & \multicolumn{1}{c|}{0.985}          & 1.263          \\ \cline{2-6} 
			& 192                                           & \multicolumn{1}{c|}{\textbf{0.262}} & \textbf{0.445} & \multicolumn{1}{c|}{0.347}          & 0.453          \\ \cline{2-6} 
			& 336                                           & \multicolumn{1}{c|}{\textbf{0.293}} & \textbf{0.422} & \multicolumn{1}{c|}{0.455}          & 0.513          \\ \cline{2-6} 
			& 720                                           & \multicolumn{1}{c|}{\textbf{1.307}} & \textbf{0.758}          & \multicolumn{1}{c|}{1.620}          & 1.787          \\ \hline
		\end{tabular}%
\end{table}

\begin{table}[ht!]
	\centering
	\caption{The simplification of models' architecture and its reason. Referring to \textit{Attention}, we mean the model developed by ~\cite{vaswani2017attention}.}
	\label{tab:simplification}
		\begin{tabular}{|l|c|c|l|}
			\hline
			Model      & \multicolumn{1}{l|}{\# encoder} & \multicolumn{1}{l|}{\# decoder} & Reason of simplification                                                                              \\ \hline
			Attention  & 6                               & 6                               & \begin{tabular}[c]{@{}l@{}}Replacement of recurrent layers \\ with encoder-decoder pairs\end{tabular} \\ \hline
			Informer   & 4                               & 2                               & \begin{tabular}[c]{@{}l@{}}Introduction of Probsparse\\ self-attention\end{tabular}                   \\ \hline
			Autoformer & 2                               & 1                               & \begin{tabular}[c]{@{}l@{}}Introduction of auto-correlation\\ mechanism\end{tabular}                  \\ \hline
			Our model  & 1                               & 1                               & \begin{tabular}[c]{@{}l@{}}Introduction of saliency \\ detection mechanism\end{tabular}               \\ \hline
		\end{tabular}%
\end{table}

\subsection{Architecture Variations}
In this ablation, we test several variations of TSD architecture and its performance in three alternative layouts: $\{3,4,5\}$ conv-down-up blocks.
We believe that the advantages of TSD architecture are rooted in its flexibility in multi-block design for a specific dataset.
Table \ref{tab:variations} compares TSD alternatives qualitatively.
All hyperparameters are the same, excluding the block design.
Most importantly, the best model design is to have a balance between forecasting accuracy and computation costs.
Therefore, in this paper, we chose a TSD architecture with four conv-down-up blocks for all experiments.

\begin{table}[ht!]
	\caption{Evaluation of architectural variations of our proposed model. The best scores are in \textbf{bold}.}
	\label{tab:variations}
	\centering
		\begin{tabular}{|l|l|cc|cc|cc|}
			\hline
			\multirow{2}{*}{Model}    & \multirow{2}{*}{Horizon} & \multicolumn{2}{c|}{\begin{tabular}[c]{@{}c@{}}3 conv-down-up\\ blocks\end{tabular}} & \multicolumn{2}{c|}{\begin{tabular}[c]{@{}c@{}}4 conv-down-up\\ blocks\end{tabular}} & \multicolumn{2}{c|}{\begin{tabular}[c]{@{}c@{}}5 conv-down-up\\ blocks\end{tabular}} \\ \cline{3-8} 
			&                          & \multicolumn{1}{c|}{MSE}                            & MAE                           & \multicolumn{1}{c|}{MSE}                            & MAE                           & \multicolumn{1}{c|}{MSE}                                & MAE                       \\ \hline
			\multirow{4}{*}{ILI}      & 24                       & \multicolumn{1}{c|}{1.519}                          & \textbf{1.081}                & \multicolumn{1}{c|}{\textbf{1.514}}                 & 1.103                         & \multicolumn{1}{c|}{1.537}                              & 1.107                     \\ \cline{2-8} 
			& 36                       & \multicolumn{1}{c|}{\textbf{1.396}}                 & \textbf{1.080}                & \multicolumn{1}{c|}{1.449}                          & 1.086                         & \multicolumn{1}{c|}{1.405}                              & 1.128                     \\ \cline{2-8} 
			& 48                       & \multicolumn{1}{c|}{1.224}                          & \textbf{0.958}                & \multicolumn{1}{c|}{\textbf{1.186}}                 & 0.971                         & \multicolumn{1}{c|}{1.278}                              & 0.978                     \\ \cline{2-8} 
			& 60                       & \multicolumn{1}{c|}{1.126}                          & \textbf{0.924}                & \multicolumn{1}{c|}{1.140}                          & 0.946                         & \multicolumn{1}{c|}{\textbf{1.125}}                     & 0.930                     \\ \hline
			\multirow{4}{*}{Exchange} & 96                       & \multicolumn{1}{c|}{0.643}                          & 1.442                         & \multicolumn{1}{c|}{\textbf{0.184}}                 & \textbf{0.369}                & \multicolumn{1}{c|}{3.109}                              & 1.406                     \\ \cline{2-8} 
			& 192                      & \multicolumn{1}{c|}{1.818}                          & 1.676                         & \multicolumn{1}{c|}{\textbf{0.262}}                 & \textbf{0.445}                & \multicolumn{1}{c|}{0.287}                              & 0.458                     \\ \cline{2-8} 
			& 336                      & \multicolumn{1}{c|}{0.934}                          & 1.343                         & \multicolumn{1}{c|}{\textbf{0.293}}                 & \textbf{0.422}                & \multicolumn{1}{c|}{0.463}                              & 0.494                     \\ \cline{2-8} 
			& 720                      & \multicolumn{1}{c|}{1.411}                          & 0.937                         & \multicolumn{1}{c|}{\textbf{1.307}}                 & \textbf{0.758}                & \multicolumn{1}{c|}{1.505}                              & 1.040                     \\ \hline
		\end{tabular}%
\end{table}


\section{Future Work}


Although the concept of TSD demonstrates remarkable experimental outcomes, it requires a thorough hyperparameter search, as outlined in Section \ref{subsec:experimental-setup}. 
The pre-defined range of hyperparameters heavily relies on domain expertise provided by an expert in the field.
Specifically, the investigation involves determining the appropriate layer size in the convolutional blocks, establishing the optimal sequence of layers and examining its impact on prediction, and exploring potential interactions between the number of heads and conv-down-up blocks. This process entails over 700 manual trial-and-error iterations in selecting the hyperparameter ranges, further enhancing performance.
We acknowledge that hyperparameter optimization remains a limitation of the model.

While U-Net architectures are predominantly employed in biomedical image segmentation to automate identifying and detecting target regions or sub-regions, our study demonstrates that TSD can be effectively utilized for time series data. It is essential to highlight that a thorough analysis of U-Net variants \cite{siddique2021u,punn2022modality}, encompassing inter-modality and intra-modality categorization, holds significance in gaining deeper insights into the challenges associated with time series forecasting and saliency detection. 
This analysis recommends a valuable avenue for future research endeavors in this field.


This study scrutinizes the efficacy of automatically encoding saliency-related temporal patterns by establishing connections with appropriate attention heads by incorporating information contracting and expanding blocks inspired by the U-Net style architecture.
To substantiate our claims, we utilize an embarrassingly simple TSD forecasting baseline. 
It is crucial to note that the contributions of this work do not solely lie in proposing a state-of-the-art model; instead, they stem from posing a significant question, presenting surprising comparisons, and elucidating the effectiveness of TSD, as asserted in existing literature, from various perspectives.
Our comprehensive investigations will prove advantageous for future research endeavors in this domain. The initial TSD model exhibits limited capacity, and its primary purpose is to serve as a straightforward yet competitive baseline with robust interpretability for subsequent research. 
TSD sets a new baseline for all pursuing multi-discipline work on popular time series benchmarks.

\section{Conclusion}
%

This research introduces an effective time series forecasting model called Temporal Saliency Detection, inspired by advancements in machine translation and down-and-up samplings in the context of image segmentation tasks. 
The proposed TSD model leverages the U-Net architecture and demonstrates superior predictability compared to existing models. 
This work's fundamental premise is the need for a technique to encode saliency-related temporal patterns through appropriate attention head connections automatically. 
The outcomes of our study underscore the significance of automatically learning saliency patterns from time series data, as the proposed TSD model significantly outperforms several state-of-the-art approaches and benchmark methods. 
In the multivariate scenario, our experiments reveal that TSD achieves a remarkable 31\% reduction in MSE and MAE losses across 46 reported results encompassing ten diverse forecasting horizons. 
Similarly, in the univariate forecasting setting, TSD yields a noteworthy 46\% reduction in both MSE and MAE losses. 
The authors conducted an ablation analysis to ensure the proposed model's effectiveness and expected behavior. 
However, it is worth noting that further potential for improvement is evident, suggesting the need for hyperparameter optimization and the exploitation of U-net variants. 
These findings underscore the efficacy and promise of the TSD model in the context of time series forecasting while also highlighting avenues for future research and refinement.

\section*{Acknowledgements}
This work is supported by the German Research Foundation (DFG) under the COSMO project (grant No. 453130567), and by the European Union's Horizon WINDERA under the grant agreement No. 101079214 (AIoTwin), and RIA research and innovation programme under the grant agreementNo. 101092908 (SmartEdge).

%
%
%
\bibliographystyle{splncs04}
\bibliography{ts2023}
%
%
%
%
%
\end{document}